\documentclass[runningheads]{llncs}
\usepackage{graphicx}
\usepackage{natbib}
\usepackage{epsfig}
\usepackage{subcaption}
\usepackage{calc}
\usepackage{amssymb}
\usepackage{amstext}
\usepackage{amsmath}

\usepackage{amsthm}
\usepackage{multicol}
\usepackage{pslatex}
\usepackage{apalike}
\usepackage{url}
\usepackage{float}

\usepackage{booktabs}
\usepackage{natbib}
\usepackage{tabularx}
\usepackage{multicol}
\usepackage{pslatex}
\usepackage{apalike}

\begin{document}

\title{Intercategorical Label Interpolation for Emotional Face Generation with Conditional Generative Adversarial Networks}
%
%

\author{Silvan Mertes \and
Dominik Schiller \and
Florian Lingenfelser \and
Thomas Kiderle \and
Valentin Kroner \and
Lama Diab \and
Elisabeth André
}

\authorrunning{Mertes et al.}
\titlerunning{Intercategorical Label Interpolation}
%
\institute{University of Augsburg, Universitätsstraße 1, 86159 Augsburg, Germany
\email{\{firstname.secondname\}@informatik.uni-augsburg.de}\\ 
}

\maketitle              
\begin{abstract}
Generative adversarial networks offer the possibility to generate deceptively real images that are almost indistinguishable from actual photographs. Such systems however rely on the presence of large datasets to realistically replicate the corresponding domain. This is especially a problem if not only random new images are to be generated, but specific (continuous) features are to be co-modeled.
A particularly important use case in \emph{Human-Computer Interaction} (HCI) research is the generation of emotional images of human faces, which can be used for various use cases, such as the automatic generation of avatars. The problem hereby lies in the availability of training data. Most suitable datasets for this task rely on categorical emotion models and therefore feature only discrete annotation labels. This greatly hinders the learning and modeling of smooth transitions between displayed affective states.
To overcome this challenge, we explore the potential of label interpolation to enhance networks trained on categorical datasets with the ability to generate images conditioned on continuous features.
\keywords{Generative Adversarial Networks, Face Generation, Conditional GAN, Emotion Generation, Label Interpolation.}
\end{abstract}

\section{Introduction}

\label{sec:introduction}

With recent advances in the field of \emph{Generative Adversarial Learning}, a variety of new algorithms have emerged to address artificial image data generation. 
The state of the art \emph{Generative Adversarial Networks} (GANs) are characterized by high image quality of generated results compared to other generative approaches such as \emph{Variational Autoencoders}. 
However, early GAN architectures lack the ability to generate new data in a controllable way. 
The original GAN framework has been modified and extended in a variety of ways in order to enable such a controlled generation of new images. 
These modified architectures have demonstrated the ability to address a broad range of image generation tasks.
Especially in the field of \emph{Human-Computer Interaction} (HCI), these systems are a promising tool.
One particularly relevant task is the generation of avatar images, which are images of human faces that can be controlled with respect to various human-interpretable features. 
In the context of emotional face generation, this enables the generation of avatar images conditioned on a particular emotion.

Most datasets suitable for training such face generation GANs refer to categorical emotion models, i.e., they contain emotion labels that were annotated in a discrete way, e.g., the emotions refer to emotional states like happy or sad. However, for many real-world use cases, such as emotional virtual agents, corresponding face images need to be generated in a more detailed manner to improve the credibility and anthropomorphism of human-like avatars. This is especially of interest during the design stage of such virtual agents, as the consistency of different modalities of virtual agents is of great importance \citep{gong2007talking, mertes2021voice} and fine-grained degrees of expressivity can enhance the perception of certain affective states of the agent, influencing (among others) the perception of the agent's personality \citep{ kiderle2021personality}.



Moreover, images that only show single emotions are not realistic in situations where smooth transitions between different emotional states are required.
Other use cases include automatically creating textures for virtual crowd generation or augmenting data for emotion recognition tasks. 
Especially in the latter case, there is a huge need for artificially created data, since continuous emotion recognition relies on non-categorical training data, and available datasets labeled in terms of dimensional features are scarce. 
In all these cases, the use of dimensional emotion models would be more sufficient to meet the requirements posed.

In this work, we explore the applicability of label interpolation for \emph{Conditional GANs} (cGANs) that were trained on categorical datasets. 
By doing so, we study the possibility to bypass the need for continuously labeled datasets. Since categorical labels are essentially binned version of continuous labels, it makes sense that the samples belonging to a specific categorical label are covering a large spectrum of expressiveness. We believe that this information can be learned by a generative model and being exploited to create emotional images on a continuous scale.
To explore the feasibility of our hypothesis, we first train cGANs on two datasets widely used for benchmarking various deep learning tasks, namely CIFAR-10 \citep{krizhevsky2009learning} and Fashion-MNIST \citep{fashion-mnist}. 
Those datasets contain discrete class labels that we use for conditioning the GAN. 
We then examine the effects of interpolating between those discrete class labels, by observing how a pre-trained classifier behaves when looking at continuously interpolated results. 
From the insights gained from these more generic datasets, we tackle the concrete use-case of emotional face generation as already described in \cite{mertes2021continuous}.
By doing so, we enable the cGAN to generate faces showing emotional expressions that can be controlled in a continuous, dimensional way. 
The goal of this paper is therefore to answer the question of whether label interpolation can be a tool to overcome the drawbacks of categorical datasets for emotional face generation.

Extending our already published work \citep{mertes2021continuous}, this paper explores not only on the applicability of label interpolation to the scenario of emotional face generation, but additionally reports on preceding experiments (see Section \ref{sec:feasibility_comp_eval}), gaining more in-depth insights into the feasibility of label interpolation.

\section{Background}


When Generative Adversarial Networks (GANs) were first introduced by Goodfellow et al. \citet{goodfellow2014generative} they sparked a plethora of research innovations in the field of artificial data generation.
GANs are based on the idea of two neural networks competing against each other in a min-max game.
While the \emph{Generator} network aims to generate new data that has never been observed before and imitates the original training domain as closely as possible, the \emph{Discriminator} tries to discern original samples from the target domain and fake sample created by the generator.
As a result, the generator learns to produce artificial data samples resembling the original domain from a random noise vector. 
Over the last years, this basic concept has been refined and advanced considerably, aiming to improve the training and output performance of GANs in various ways \citep{arjovsky2017towards, gulrajani2017improved, arjovsky2017wasserstein}. 
Along with setting the new state-of-the-art of quality for artificial image generation, GANs have opened up new possibilities for facial image generation and modification.\\

\citet{radford2015unsupervised}, for instance, modified the original GAN architecture by exchanging the fully-connected layer architecture with convolutional networks in the generator and discriminator, allowing, among others, to generate high quality human face images. 
Additionally, they investigated how their so-called \emph{Deep Convolutional GAN} (DCGAN) implicitly maps the latent space to facial features (e.g. face pose). However, since the  training procedure of DCGAN is unsupervised, it comes with the inherent drawback that the facial feature types to be learned can not be directly controlled.
\citet{karras2017progressive} presented \emph{Progressive Growing GANs}, an even more sophisticated approach to adversarial image generation, that they applied to the task of human face generation.\\

%
The aforementioned approaches all underlay the drawback that their outputs solely depend on the random noise input vector, without the possibility to control it in a human readable way.
This problem was addressed by  \citet{mirza2014conditional}, who use \emph{Conditional GANs} (cGAN) to encode  additional label information in the input vector, enabling the network to consider certain pre-defined features in the output.
This property of cGANs was exploited by \citet{wang2018attributes} and \citet{gauthier2014conditional} to generate face image data with respect to specific features (e.g. \emph{glasses}, \emph{gender}, \emph{age}, \emph{mouth openness}).
Similarly, \citet{yi2018data} made use of the cGAN conditioning mechanisms in order to augment emotional face image datasets. 
One problem of this approach is the usage of either using discretely labeled features, restricting the output to discrete categories, or to already use continuously labeled data during training which is rarely available in a plethora of scenarios.

A related task that GANs are frequently applied to is the task of \emph{Style Conversion}, which in terms of facial expressions is also known as \emph{Face Editing}. 
It intends to modify existing image data instead of generating entirely new data \citep{he2019attgan, royer2020xgan, choi2018stargan, liu2017unsupervised, lin2018conditional}. 
Using GANs, \citet{ding2018exprgan}
managed to develop a framework that allows to continuously adapt the emotional expressions of images. Although their approach is not explicitly based on continuously annotated data, the diversity of the intensity of emotions must be represented in the training set. Their system proved its capability of generating random new faces expressing a particular emotion. However, they didn't investigate the generation capabilities of their system according to common known dimensional emotion models like Russel's Valence-Arousal model \citep{russell1999core}. The focus was rather to show that their face editing system is able to modify the intensity of discrete, categorical emotions.

In general, interpolating through the label space of a cGAN is a quite under-explored mechanism. Direct manipulation of the latent input space of GANs has been made possible by various automated approaches like \emph{Latent Vector Evolution} \citep{schrum2020interactive} as well as interactive ones \citep{schlagowski2021taming}. Also, in generative approaches apart from adversarial learning, exploring interpretable and non-interpretable latent spaces are a widespread tool, for example in the context of Human-Robot Interaction \citep{ritschel2019personalized} or Speech Synthesis \citep{rijn21_interspeech}. However, manipulating the discrete label space of cGANs in a continuous way has not yet found its way into practice.
To the best of the authors' knowledge, there is no system that is trained on discrete emotion labels and outputs new face images that can be controlled in a continuous way.

\section{Technical Framework for interpolating Categorical Labels}
In order to explore the applicability of label interpolation in cGANs, an appropriate framework had to be defined, which is presented in the following sections.

\subsection{Network Architecture}\label{architecture}
The networks utilized in our experimental settings are largely founded on a \emph{Deep Convolutional GAN} (DCGAN) by \citet{radford2015unsupervised}.
A detailed description of the original DCGAN architecture can be found in the respective publication. In summary, DCGANs are a modification to the original GAN framework by \citet{goodfellow2014generative}, where convolutional and convolutional-transpose layers were included in order to model the training domain with higher image quality. The architectures that we used were modified to fit the corresponding datasets. 
Additionally, to enable targeted image generation (which is not part of the original DCGAN), the architectures were extended with the principles of a \emph{cGAN}.

Unlike conventional GANs, cGANs incorporate a conditioning mechanism consisting of an additional class input vector. This vector is used to control specific features of the output images by telling the generator network about the presence of certain features during training. Thus, this feature information must be given as labels while training the cGAN. Thus, the input for a cGAN consists of a random noise component $z$ (as in the original GAN framework) and a conditioning vector $v$. After the training process, the generator has learned to transform the random noise input into images that resemble the training domain, taking into account the conditioning information given by $v$ in order to drive the outputs to show the desired features. 


In our implementation, the conditioning information is given to the network as one-hot encoded label vector, where each element represents a certain feature. 
Thus, the one-hot label vector $v$ has the following form \citep{mertes2021continuous}:
\begin{equation}
v = (v_1, v_2, ..., v_n) = \{0, 1\}^n
\end{equation}
where $n$ is the number of controlled features. The datasets that we used in our experiments are primarily designed for classification tasks. This implies that we consider a feature a class of the dataset. As, in the scope of this work, only datasets for single-class classification were considered, the following restriction holds true \citep{mertes2021continuous}: 
\begin{equation}\label{equation_sum_equals_one}
\sum_{i=1}^{n}v_i = 1
\end{equation}

\subsection{Interpolation}\label{sec:interpolation}

After training, the definition of the condition part of the cGAN's input vector is changed to allow for a continuous interpolation between the originally discrete classes. Generally, this can simply be done by reformulating the conditining vector $v$ so that is not forced to a binary structure \citep{mertes2021continuous}:
\begin{equation}
v = (v_1, v_2, ..., v_n) = [0,1]^n
\end{equation}
During our experiments, we found that keeping the restriction formulated in Equation \ref{equation_sum_equals_one} leads to better quality of interpolated results instead of picking the single elements of the vector arbitrarily in the interval $[0,1]$. 
In other words, interpolation is done by subtracting some portion $e$ from the input representative of one class and adding it to another class.
Our hypothesis is that due to the differentiable function that is approximated by the cGAN model during the training process, those non-binary conditioning vectors lead to image outputs which are perceived as lying somewhere \emph{between} the original, discrete classes. For our target context, the generation of face images with continuous emotional states, this would refer to images of faces that do not show the extreme, discrete emotions that are modeled in a categorical emotion system, but to more fine-grained emotional states as they are conventionally modeled by a dimensional emotion model as will be further elaborated on in Section \ref{sec:emotion}.

\section{Feasibility Studies}\label{sec:feasibility_comp_eval}
To evaluate the feasibility of our approach, we decided to first apply it to two generic datasets, before finally addressing the problem of emotional human face generation. 

\subsection{Datasets}
\label{sec:feasability_datasets}

\begin{figure*}[h]
	\centering
	\includegraphics[width=1\textwidth]{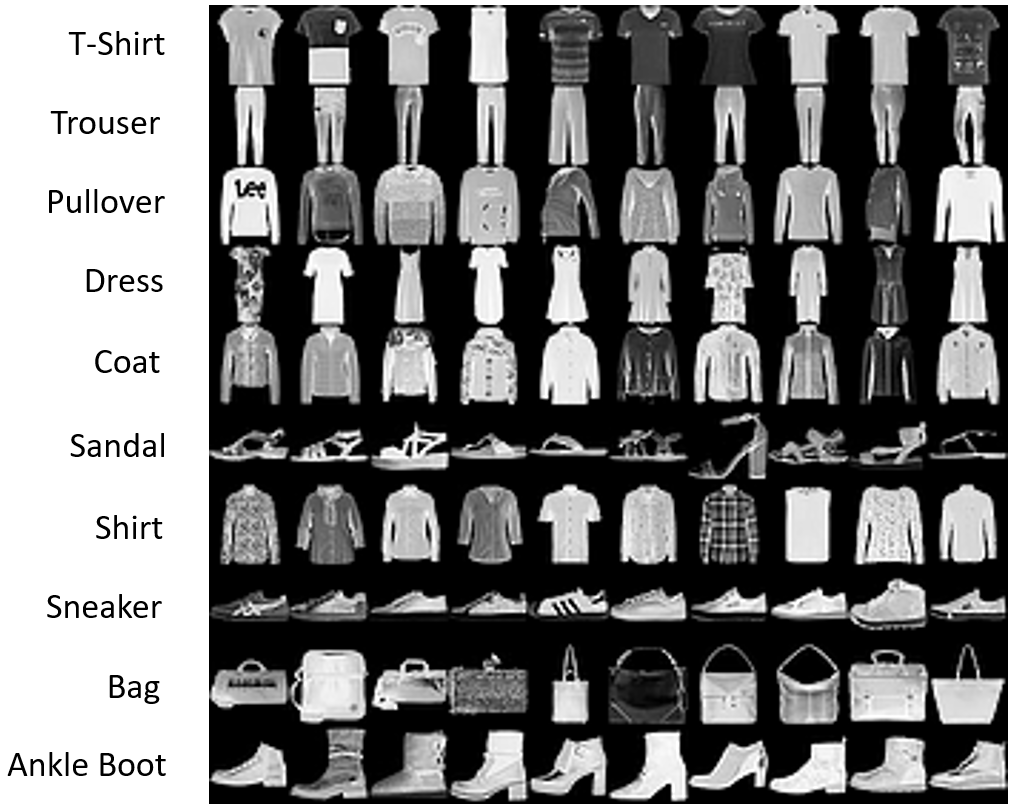}
	\caption{Fashion-MNIST categories and examples. \citep{fashion-mnist}}
	\label{fig:mnist_morph}
\end{figure*}

The Fashion-MNIST dataset \citep{fashion-mnist}  encompasses a set of product pictures taken from the Zalando website, where each image belongs to one of 10 classes. Each of these contains 7,000 pictures. The images that we used are 8-bit grayscale versions with a resolution of 28x28 pixels. All in all, this results in a dataset of 70,000 fashion product pictures, whereas 60,000 are attributed to the training dataset and 10,000 to the test set. Examples for each class are depicted by figure \ref{fig:mnist_morph}. \\

\begin{figure*}[h]
	\centering
	\includegraphics[width=1\textwidth]{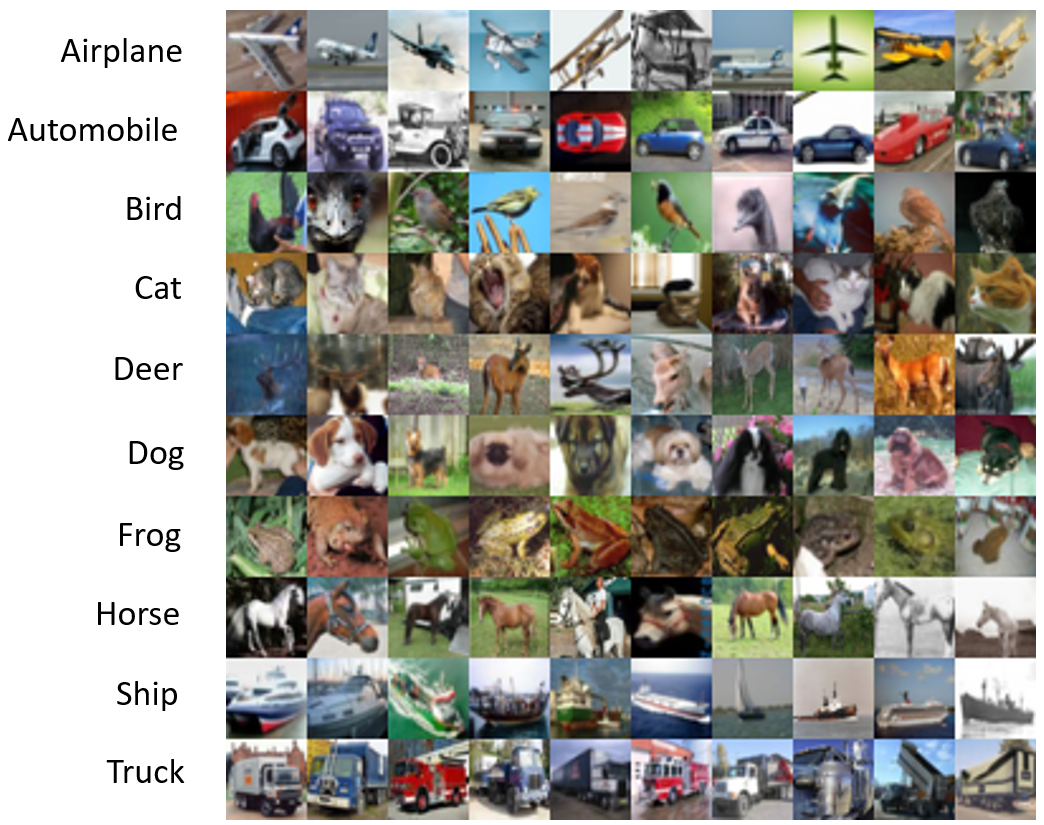}
	\caption{CIFAR-10 categories and examples. \citep{krizhevsky2009learning}}
	\label{fig:cifar_10}
\end{figure*}

The CIFAR-10 and the CIFAR-100 datasets both are derived from the \emph{80 million tiny images dataset} \citep{krizhevsky2009learning}. In contrast to the 100 classes of CIFAR-100, CIFAR-10 only contains a subset of 10 classes, whereas each class has 6,000 colored images of size $32x32$. This results in a dataset of 60,000 images in total, where 50,000 belong to the training and 10,000 to the test set. The classes are mutually exclusive, even for narrow classes like trucks and cars. Figure \ref{fig:cifar_10} depicts example images for the corresponding 10 classes.

We decided to use the Fashion-MNIST dataset because it has originally been designed for measuring the performance of machine learning approaches. The pictures are grayscaled and comparably small, making the dataset suitable for preliminary feasibility experiments.
To further test the viability of our approach, we aimed to increase the challenge gradually. Thus, we additionally chose to use the CIFAR-10 dataset. Although it also contains small pictures, the challenge is raised by the colorization and the slightly higher resolution.

\subsection{Methodology}
In order to evaluate if the interpolation algorithm creates smooth transitions between two arbitrary classes, we decided to perform a fine-grained analysis on the continuously generated outputs by the use of our approach. To this end, we used pre-trained classifiers that are able to accurately distinguish between the different discrete classes contained in the respective datasets. As the focus of this work is to gain insights into the question whether interpolating between discrete label information can be a promising tool for future applications, the discrete decision of such classification models are not a good metric for our purposes. Instead, we want to explore if the interpolation mechanism is able to model the full bandwidth of transitional states that can occur \emph{between} different classes. Thus, for evaluating if the interpolation mechanism works correctly, we assessed the confidence of the classification models that the interpolated result belongs to certain classes. Ideally, during interpolation, this confidence should continuously shift towards the class that is interpolated to.

\subsection{Training}

\begin{figure}
\centering
 	\includegraphics[width=0.8\textwidth]{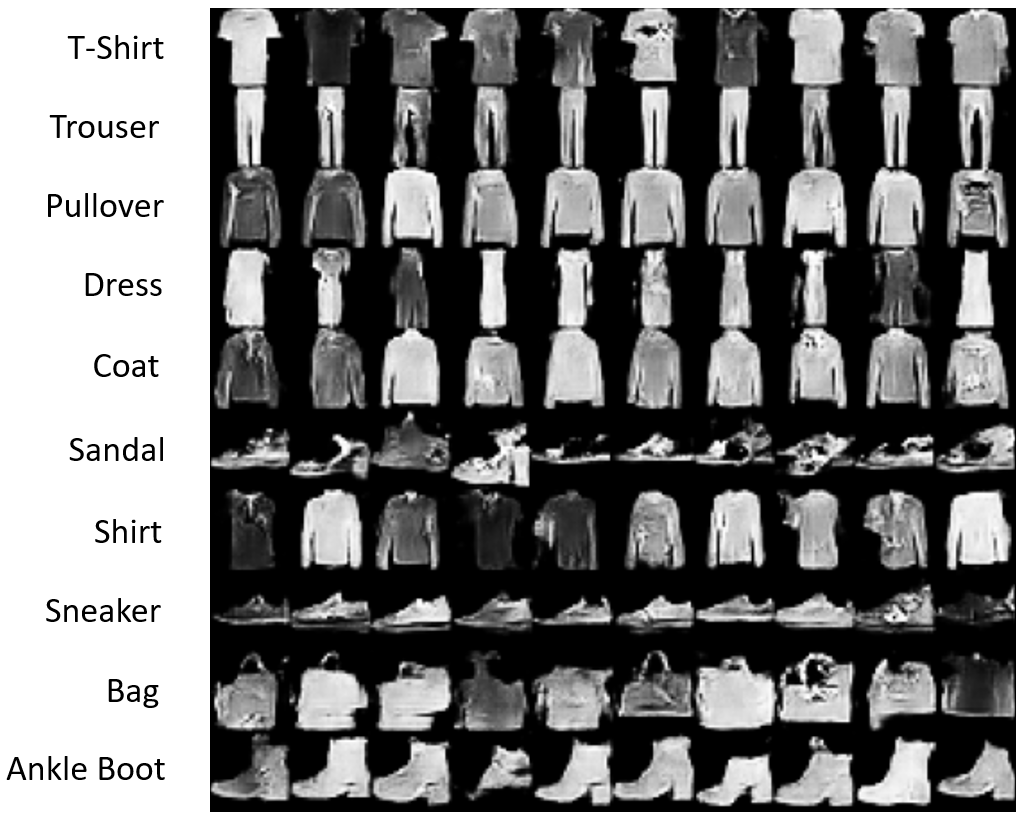}
	\caption{Exemplary outputs of the cGAN model trained on Fashion-MNIST.}
	\label{fig:fashin_examples}
\end{figure}
\begin{figure}
	\centering
	 	\includegraphics[width=0.8\textwidth]{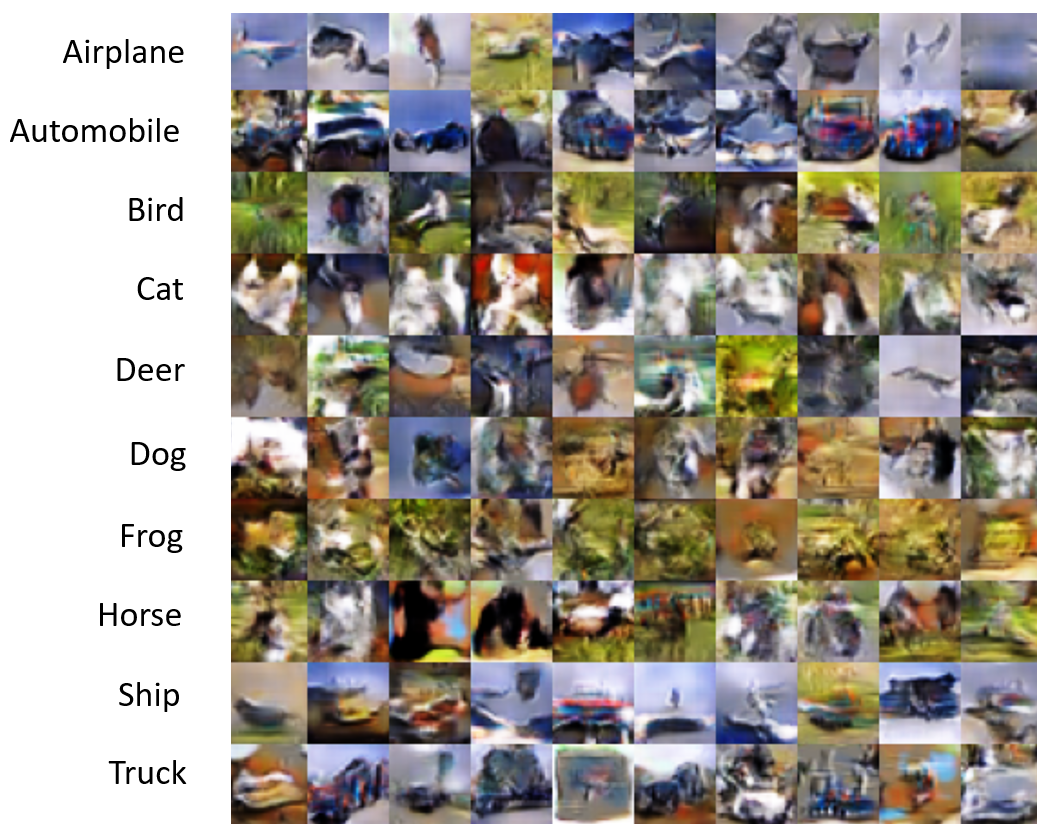}
	\caption{Exemplary outputs of the cGAN model trained on CIFAR-10.}
	\label{fig:cifar_examples}

\end{figure}

\begin{figure}
\centering
 	\includegraphics[width=0.8\textwidth]{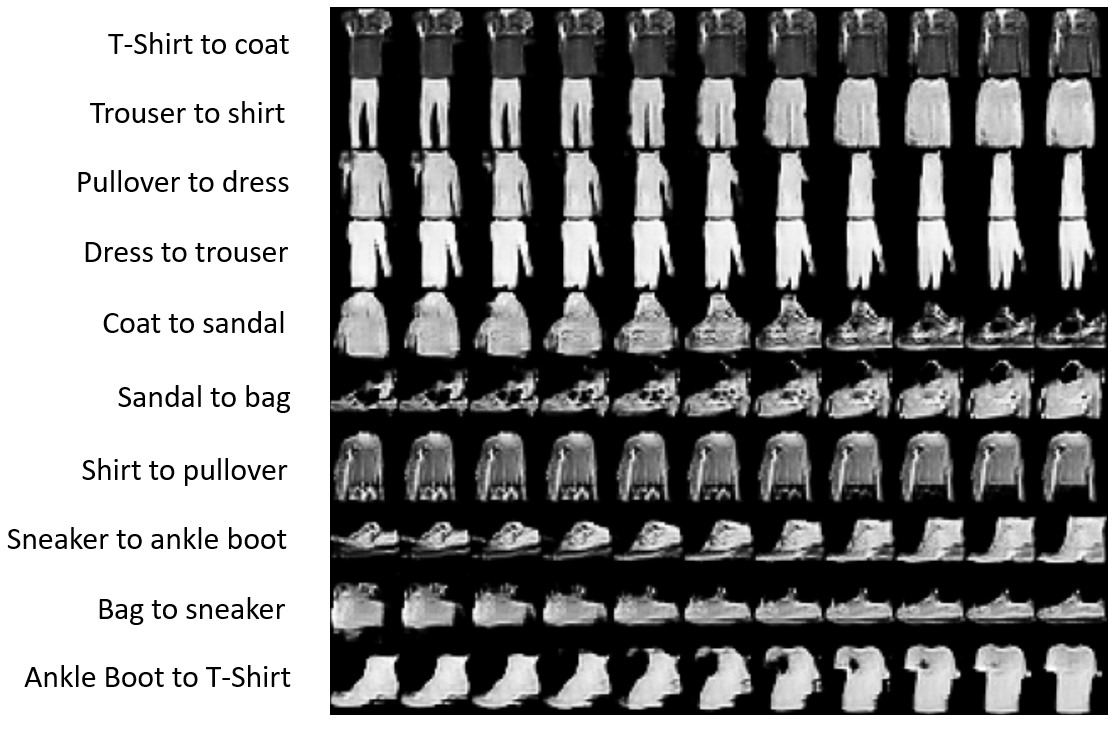}
	\caption{Exemplary outputs of the interpolation steps of the cGAN model trained on Fashion-MNIST.}
	\label{fig:fashion_morphs}
\end{figure}
\begin{figure}
	\centering
	 	\includegraphics[width=0.8\textwidth]{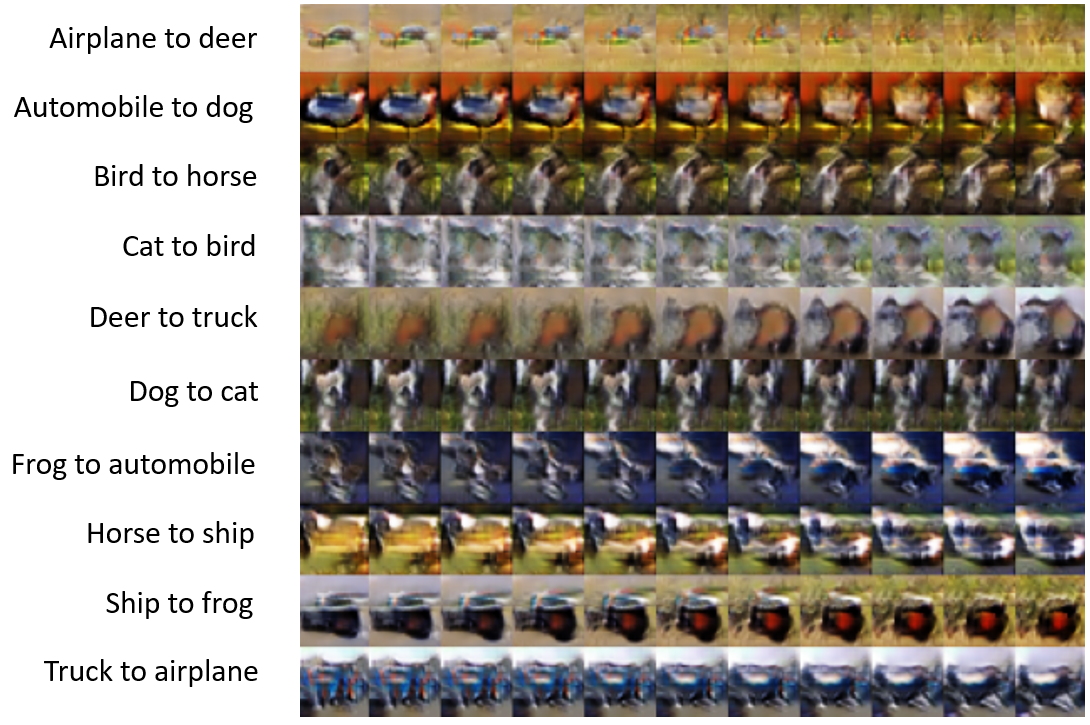}
	\caption{Exemplary outputs of the interpolation steps of the cGAN model trained on CIFAR-10.}
	\label{fig:cifar_morphs}

\end{figure}

For both the datasets, we adapted the DCGAN architecture to fit the dataset. Slight changes to the architecture had to be made in order to produce reasonable outputs. Further, we enhanced both models with the conditioning mechanism as described in Sec. \ref{architecture}. 

\emph{Fashion-MNIST.} For this dataset, we trained the cGAN model for 20,000 random batches of size 32 on all of the 50,000 images of the \emph{train} partition of the dataset using Adam optimizer with a learning rate of 0.0002 and $\beta_1$ of 0.5. Example outputs of the trained model can be seen in Fig. \ref{fig:fashin_examples}, whereas example outputs of different interpolation steps are shown in Fig. \ref{fig:fashion_morphs}.

\emph{CIFAR-10.} For this dataset, we trained the cGAN model for 30,000 random batches of size 32 on all of the 50,000 images of the \emph{train} partition of the dataset, again using Adam optimizer with a learning rate of 0.0002 and $\beta_1$ of 0.5. Example outputs of the trained model can be seen in Fig. \ref{fig:cifar_examples}, whereas example outputs of different interpolation steps are shown in Fig. \ref{fig:cifar_morphs}. In both the images, it can be clearly seen that the chosen cGAN architecture apparently was not able to resemble the traing domain sufficiently enough. Results are blurry, and objects can only partially be recognized as the intended objects. However, we chose to continue with the validation of the interpolation as we were also interested in how label interpolation behaves when dealing with models that do not represent the respective training domain very well.

\begin{figure}[htbp]
\captionsetup{justification=raggedright,margin=.1cm}
 \begin{minipage}{0.5\textwidth}
 \includegraphics[width=\textwidth]{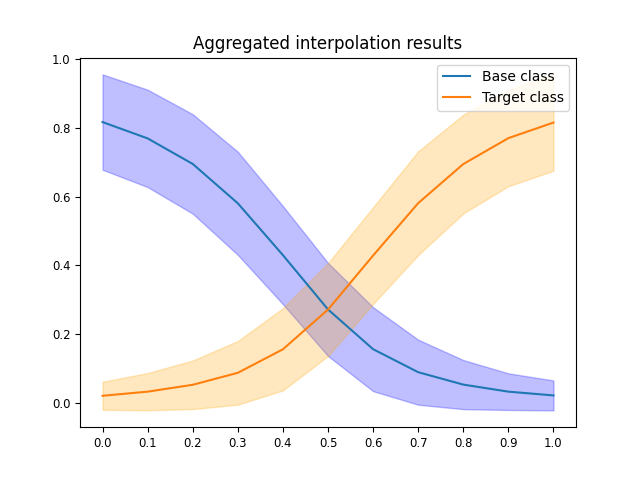}
 \caption{Results of the computational evaluation with Fashion-MNIST.}
 \label{fig:comp_eval_fashion}
 \end{minipage} 
	\hfill
	\begin{minipage}
	{0.5\textwidth}
	
	 \includegraphics[width=\textwidth]{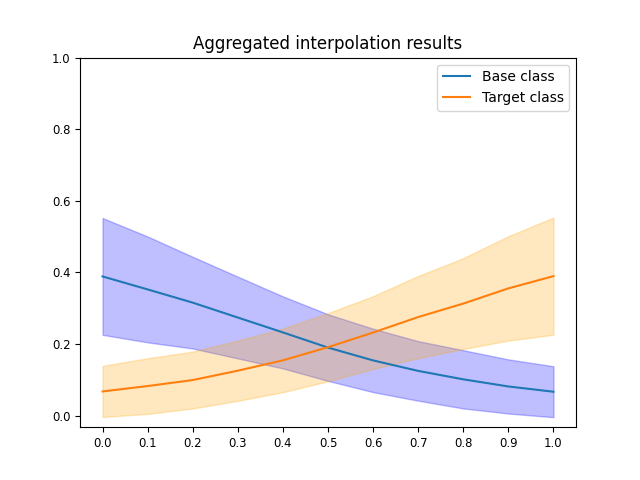}
\caption{Results of the computational evaluation with CIFAR-10.}
\label{fig:comp_eval_cifar}
 \end{minipage}
\end{figure}

\subsection{Computational Evaluation}
\label{sec:ceval}

In order to test the capability to interpolate between different classes, we used classifiers that we trained on the task of object classification. To this end, we used the EfficientNet-B0 architecture \citep{tan2019efficientnet}, as these models turned out to achieve very high accuracy on both datasets (\emph{Fashion-MNIST: $0.9089$, CIFAR-10: $0.9931$}). We used a softmax layer on top of the models, which produces an output vector \(r \in I\!R^+\ ^n\) with \(\sum_{i=1}^n r_i = 1\) where $n$ is the number of classes. By interpreting this class probability vector $r$ as confidence distribution over all the classes, we can assess the interpolation capabilities of the cGAN models by observing the change of $r$. To this end, 1,000 image sets were randomly generated for each class combination $i,j$ in CIFAR-10 as well as Fashion-MNIST. Each of these images was conditioned on the respective source class $i$. 
Then, we performed interpolation steps for every source image as described in Sec. \ref{sec:interpolation} with $\alpha = 0.1$, resulting in 10 interpolation steps until the target class was reached. For each interpolation steps, we fed all resulting images into the respective classifier model (i.e., either the Fashion-MNIST or the CIFAR-10 model).  Results of the computational evaluation are plotted in Fig. \ref{fig:comp_eval_fashion} and Fig. \ref{fig:comp_eval_cifar}. 


\section{Dimensional Face Generation}
As our feasibility studies revealed, that the mechanism of label interpolation shows promise when being used with more generic datasets, we apply it to our desired scenario of emotional face generation, as we already described in \citep{mertes2021continuous}.

\subsection{Emotion Models}\label{sec:emotion}
\begin{figure}[h]
	\centering
	\includegraphics[width=0.45\textwidth]{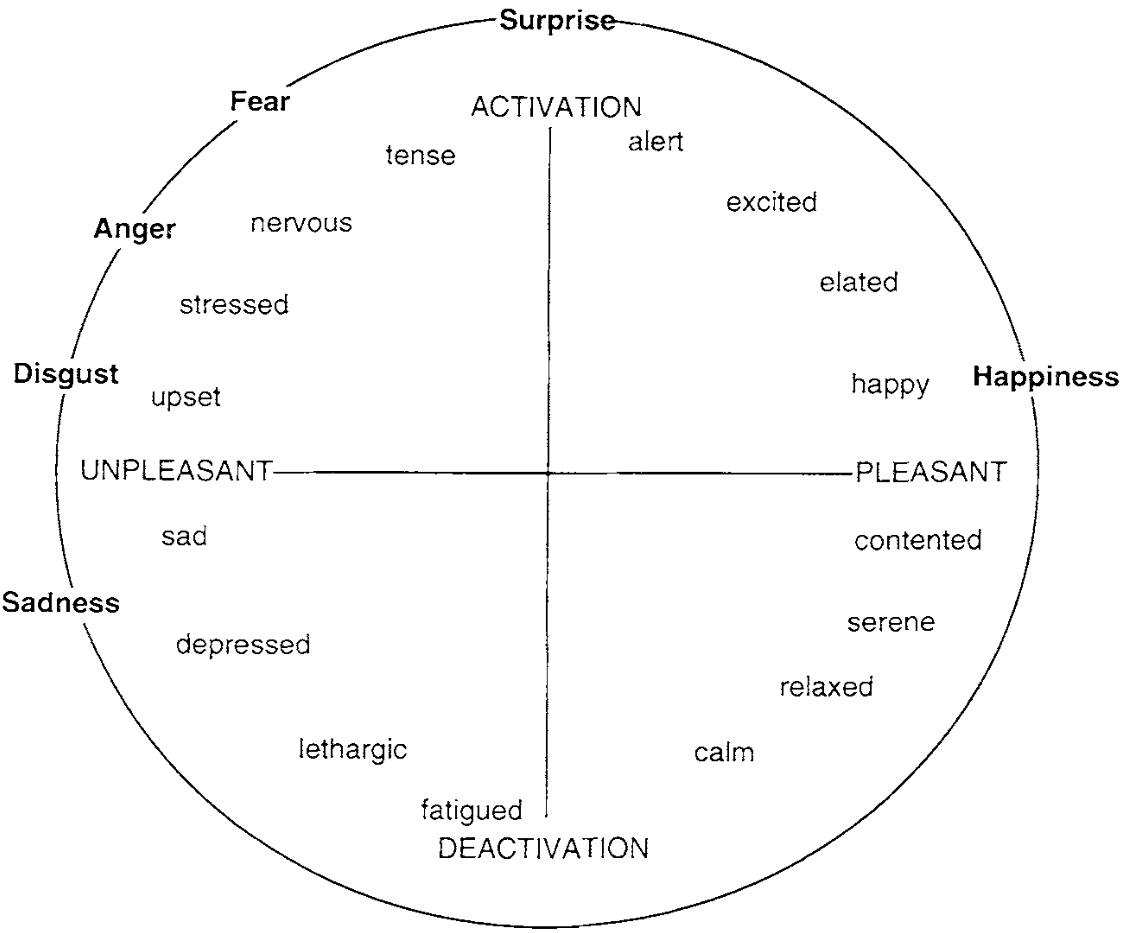}
	\caption{Russel's 2-dimensional valence arousal circumplex \citep{russell1999core}.}
	\label{fig:val_ar}
\end{figure}

Enabling algorithms to handle human emotion requires a discrete  definition of affective states. 
Categorical and dimensional models are the two most prevalent approaches to conceptualize human emotions.\newline 
A categorical emotion model subsumes emotions under discrete categories like happiness, sadness, surprise or anger. 
There is a common understanding of these emotional labels, as terms describing the emotion classes are taken from common language. 
It is also for this reason, that labels are the more common form of annotation found with datasets depicting emotional states. 
However, this (categorical) approach may be restricting, as many blended feelings and emotions cannot adequately be described by the chosen categories. 
Selection of some particular expressions can not be expected to cover a broad range of emotional states, especially not differing degrees of intensity.\newline 
An arguably more precise way of describing emotions is to attach the experienced stimuli to continuous scales within dimensional models. 
\citet{Mehrabian1995} suggests to characterize emotions along three axes, which he defines as pleasure, arousal and dominance. 
\citet{Lang1997} proposes the simplified axes of arousal and valence as measurements, resulting in the more commonly used dimensional emotion model. 
The valence scale describes the pleasantness of a given emotion. 
A positive valence value indicates an enjoyable emotion such as joy or pleasure. 
Negative values are associated with unpleasant emotions like sadness and fear. 
This designation is complemented by the arousal scale which measures the agitation level of an emotion (Figure \ref{fig:val_ar}). 
This representations is less intuitive but allows continuous blending between affective states.\newline

Categorical as well as dimensional models are simplified, synthetic descriptions of human emotions and are not able to cover all of the included aspects. 
However, with our interpolation approach we aim to cover all the whole emotional range defined within the space of the dimensional valence-arousal model and enable a seamless transition between displayed emotions.
As data collections featuring dimensional annotation for facial expressions are more sparse than the ones containing categorical labels (Section \ref{sec:faces_dataset}), being able to use emotional labels in the training process is very beneficial. 
Goal of the following study is to use a cGAN that was conditioned on categorical emotions during training, and interpolate between those emotions in order to be able to create new images. 
Those newly generated face images show emotional states that are located in the continuous dimensional space of the valence/arousal model without having to correlate directly with discrete emotion categories.

To formally represent the valence and arousal of a face image $I$, we use a tuple $VA(I) = (v, a)$, where $v$ refers to valence and $a$ to arousal.
Correlating with Russel's theory explained above, an image $x$ with $VA(x) = (0,0)$ is representing the center of the emotion space and thus show a neutral emotion. Emotions that are referred to in categorical emotion systems (e.g., \emph{Happy}, \emph{Sad}) are represented by valence/arousal states that show quite extreme values.
When it comes to the interpolation of those dimensional emotional states, i.e., to create images with certain degrees of arousal or valence, we interpolate between the \emph{neutral} emotion and the \emph{extreme} emotional states. By the term \emph{extreme emotion}, we refer to all categorical emotional states used except the \emph{neutral} state, as this represents the center of the dimensional emotion model.

In our experiments, we stuck to performing interpolations between \emph{Neutral} and a particular other emotion to preserve comparability between emotions. It should be noted that the approach could easily be extended to interpolate between two or even more categorical emotions. However, since we use only one categorical emotion and \emph{Neutral} at a time, the following restriction must be added:
\begin{equation}
\exists{i_{\in{[2,6]}}}:v_1 + v_i = 1
\end{equation}
where $v_1$ represents the condition for \emph{Neutral}.

To create an image that should show a specific degree of valence $v$ or arousal $a$, where $0 \leq a, v \leq 1$, we use the one-hot element of the emotion that maximizes the specific value, for example \emph{Happy} when it comes to valence, or \emph{Angry} for arousal, and then decrease it to the desired degree. At the same time, we increase the one-hot element related to \emph{Neutral} by the same amount, which allows us to create images showing valence/arousal values anywhere in Russel's emotion system, as opposed to the \emph{extreme} values given during training.

\subsection{Dataset}\label{sec:faces_dataset}
\begin{figure}[h]
	\centering
	\includegraphics[width=0.9\textwidth]{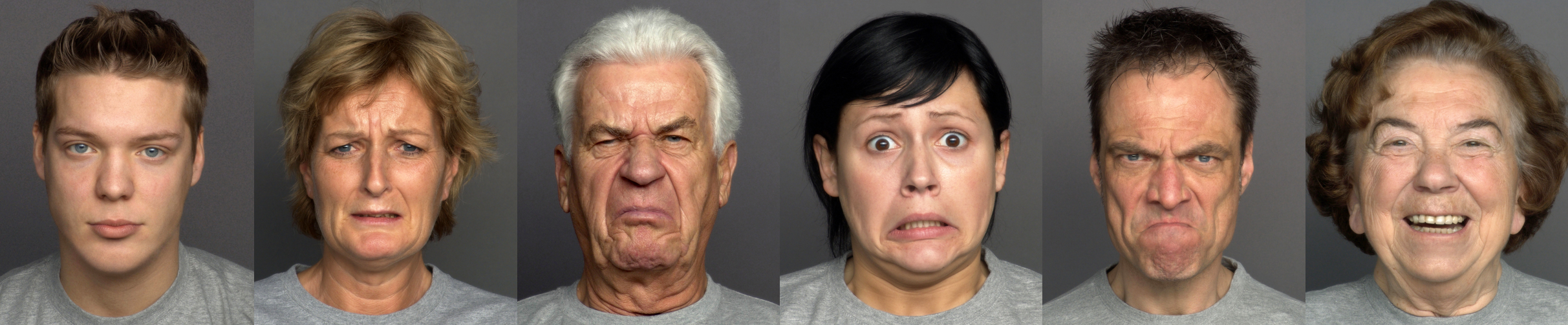}
	\caption{Exemplary data from FACES showing neutral, sad, disgust, fear, anger and happiness from left to right varying the age group. \citep{faces}}
	\label{fig:face_example}
\end{figure}

As previously mentioned, datasets labeled in terms of dimensional emotional models are scarce. Although there are a few datasets with continuous labeled information (e.g. \emph{AffectNet} by \citet{mollahosseini2017affectnet} or AFEW-VA by \citet{kossaifi2017afew}), they use to be gathered in the wild, resulting in miscellaneous data. 

Data diversity usually is beneficial for deep learning tasks, however, in our specific use case of face generation with the focus on modeling certain emotional states in human faces, consistency in all non-relevant characteristics (i.e., characteristics not related to facial expressivity) is an advantage.

Thus, although a variety of categorically labeled datasets are available \citep{ck+,jacfee,msfde,iaps,adfes,nimstim}, we decided to use the FACES dataset \citep{faces} for our experiments, since it meets our requirements particularly well. In this dataset all images are labeled in a discrete manner, and recorded with an identical uniformly coloured background and an identical grey shirt. This is exemplified in Fig. \ref{fig:face_example}.
To overcome the disadvantages of continuously labeled, but inconsistently recorded emotional face datasets, we explore the use of label interpolation with categorically labeled datasets.


Overall the FACES dataset consists of 2052 emotional facial expression images, distributed over 171 men and women. The 58 participants are assigned to the group young, 56 to middle-aged and 57 to the old group, each showing 2 styles of the emotions \emph{Neutral}, \emph{Fear}, \emph{Anger}, \emph{Sadness}, \emph{Disgust} and \emph{Happiness}. For training we only needed to resize the pictures to a target resolution of 256x256 pixels.

\subsection{Methodology}

As our feasibility study revealed, the interpolation approach has potential for creating transitions between different discrete states. However, it could be seen that the quality of the generated images, especially when dealing with the CIFAR-10 dataset, left room for improvement. To use the approach of label interpolation in a real world scenario like avatar generation or similar, such a poor image quality would be unacceptable. Thus, besides optimizing the cGAN model for our face generation use case even more, our evaluation process here is two-folded. First, we evaluate whether the cGAN is, before applying any interpolation, able to create images that are perceived correctly by human judgers. By doing so we can assess if the cGAN model that we trained is capable of generating images with sufficient enough quality to express emotional states. Secondly, we conducted a computational evaluation analogously to the feasibility study.

\begin{figure*}[h]
	\centering
	\includegraphics[width=1\textwidth]{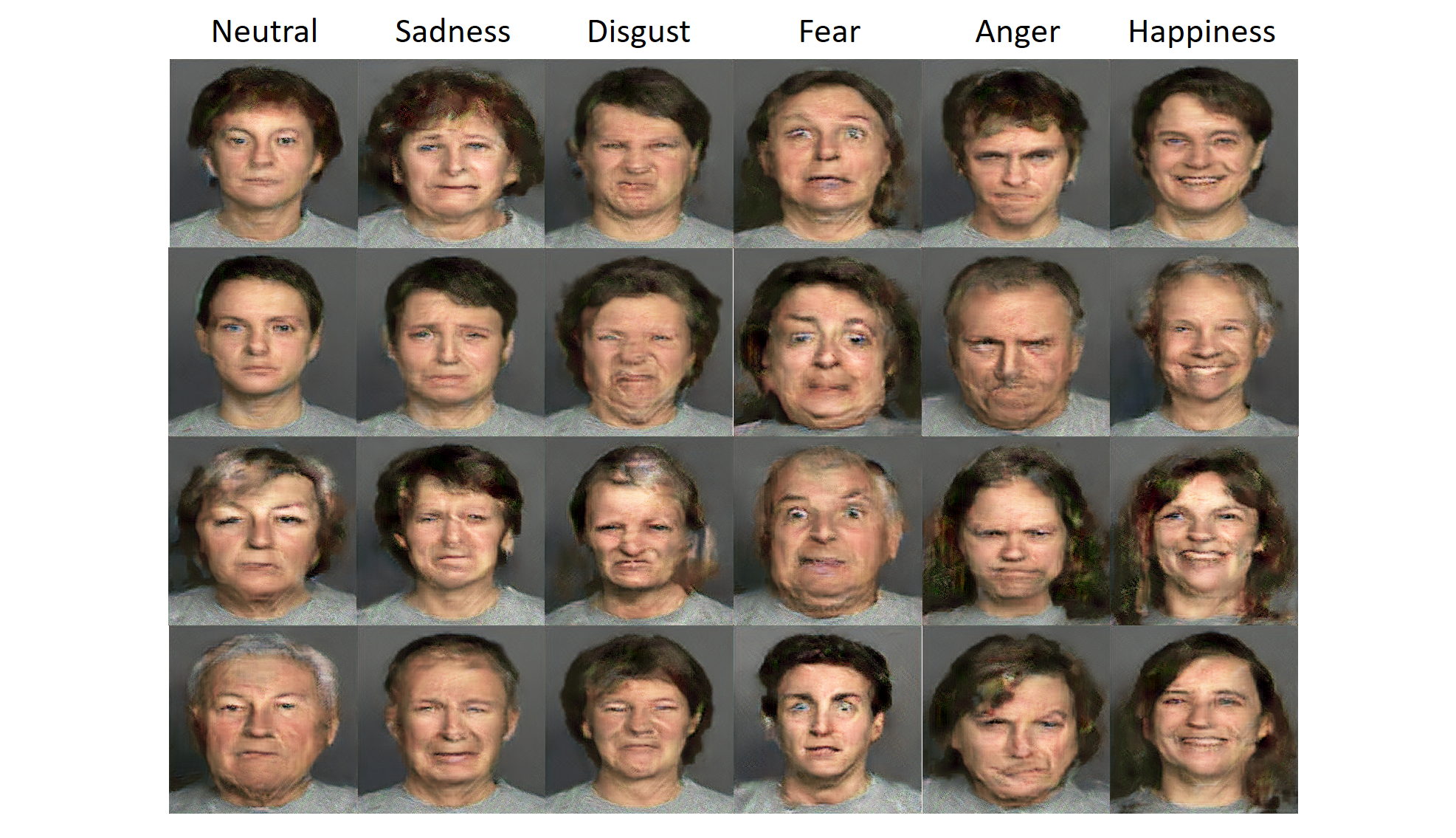}
	\caption{Example outputs of the trained cGAN model. \cite{mertes2021continuous}}
	\label{fig:cgan_discrete_samples}
\end{figure*}

\subsection{Training}
The model was trained for 10,000 epochs on all 2052 images of the FACES dataset using \emph{Adam} optimizer with a learning rate of 0.0001. Example outputs of the trained model, conditioned on one-hot vectors of all 6 used emotions, are shown in Fig. \ref{fig:cgan_discrete_samples}.

\begin{figure}[]
\minipage{0.32\textwidth}
  \includegraphics[width=\linewidth]{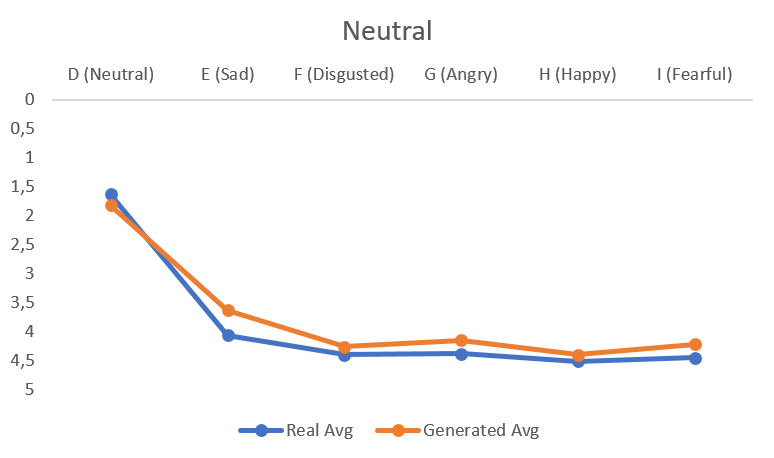}
\endminipage
\minipage{0.32\textwidth}
  \includegraphics[width=\linewidth]{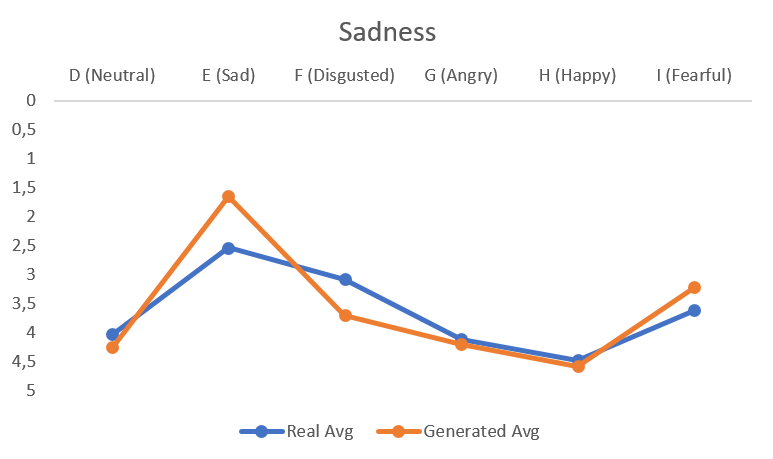}
\endminipage
\minipage{0.32\textwidth}%
  \includegraphics[width=\linewidth]{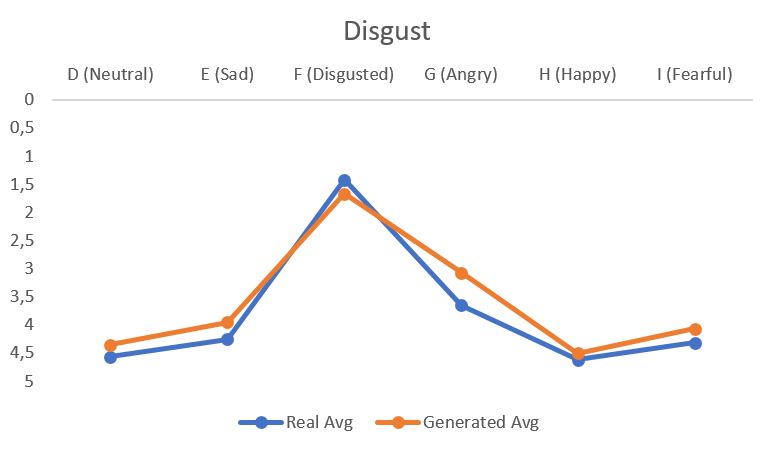}

\endminipage\hfill
\minipage{0.32\textwidth}%
  \includegraphics[width=\linewidth]{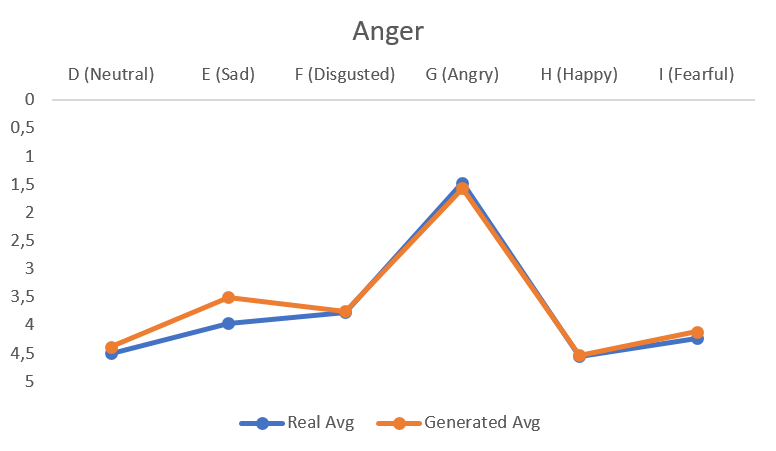}

\endminipage
\minipage{0.32\textwidth}%
  \includegraphics[width=\linewidth]{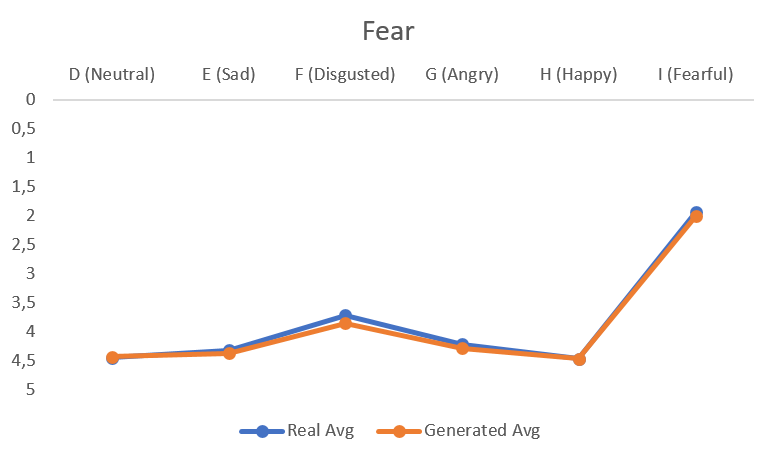}
\endminipage
\minipage{0.32\textwidth}%
  \includegraphics[width=\linewidth]{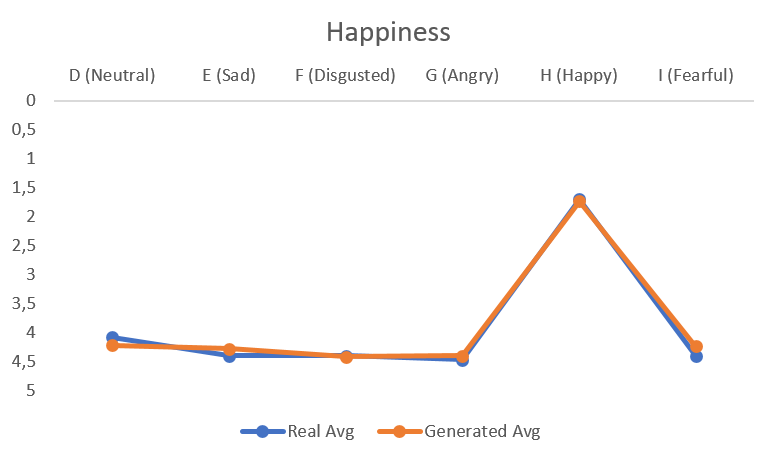}
\endminipage
\caption{Results of the user study. Blue graphs show the perceived emotion of real images from the FACES dataset, while orange graphs show the perceived emotion of outputs of the cGAN conditioned on one-hot vectors. The y-axis represents the degree of the participant's agreement with the corresponding emotions that are represented by the x-axis. \citep{mertes2021continuous}}
\label{fig:eval_study}
\end{figure}

\subsection{User Evaluation}\label{sec:user_eval}
In our user study, we evaluated the cGAN's ability to produce images of discrete emotions generated with the respective one-hot vector encoding.
In total, 20 probands of ages ranging from 22 to 31 years (M = 25.8, SD = 2,46, 40\% male, 60\% female) participated in the study. 

During the survey, 36 images were shown to each of the participants. 18 of the images where original images taken from the FACES dataset, whereas the other 18 images were generated by the trained cGAN. All images were split evenly between all emotions, both for the original as well as for the generated images. To keep consistency with the images generated by the cGAN, the images taken from the FACES dataset were resized to $256x256$ pixels.
For each image, the participants were asked how much they agreed to the image showing a certain emotion. To mitigate confirmation bias, they were not told which emotion the image should show, but asked to provide their rating for each emotion. The ratings were collected by the use of a 5-point Likert scale (1 = strongly agree, 5 = strongly disagree). 
Results of the user study are shown in Fig. \ref{fig:eval_study}. 

As can be seen, the images that were generated by the cGAN were rated to show the respective targeted emotion in a similar convincing way as the original images taken from the FACES dataset.
Each emotion is mostly recognized in the correct way by the study participants.
One emotion, namely \emph{Sadness}, even stands out as the artificially generated images were recognized even better than the original images, which were mistaken for \emph{Disgust} more frequently.
Considering these results, the trained cGAN model proves to be an appropriate basis for interpolation experiments.

\subsection{Computational Evaluation}

Analogously to the computational evaluation in our feasibility studies, we verified if label interpolation can be used to enhance the cGAN network with the ability to generate images with continuous degrees of valence and arousal with the help of an auxiliary classifier.
Again, 1,000 noise vectors per class were initially fed into the cGAN, where here, the classes were the five emotions \emph{Sadness}, \emph{Disgust}, \emph{Fear}, \emph{Anger} and \emph{Happiness}.
The conditioning vector was initially chosen to represent the neutral emotion.
For each of the 5,000 noise vectors, 10 interpolation steps with step size $e=0.1$ towards the respective extreme emotion were conducted. Thus, the last interpolation step results in a one-hot vector representing the respective extreme emotion. 
For evaluting the resulting valence/arousal values, we again used a pre-trained auxiliary classifier.

\begin{figure*}[]
	\centering
	\includegraphics[width=1\textwidth]{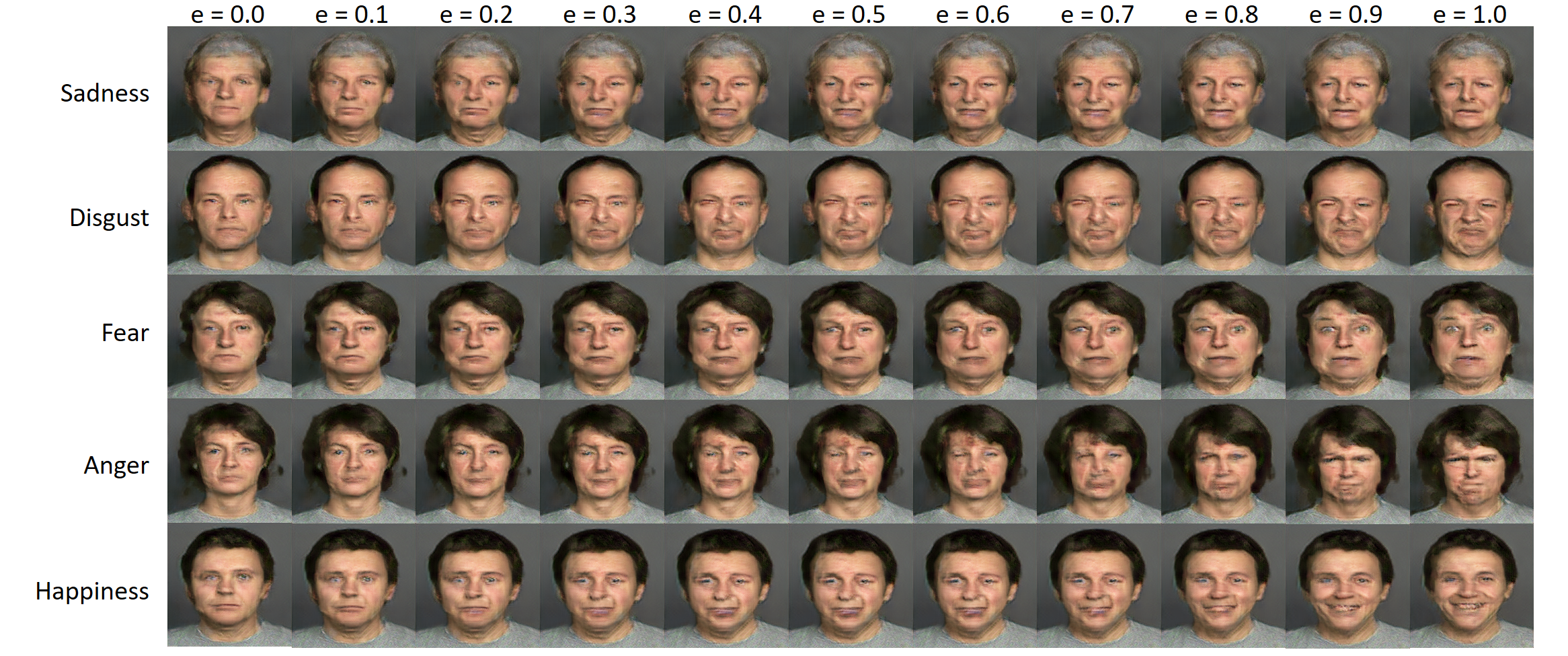}
	\caption{Example outputs of the interpolation mechanism. Each row shows a set of interpolation steps, where in each step, the emotion portion $e$ was increased by 0.1, whereas the neutral portion was decreased by the same amount. \cite{mertes2021continuous}}
	\label{fig:cgan_interpolation_samples}
\end{figure*}

Fig. \ref{fig:cgan_interpolation_samples} shows exemplary outputs of interpolation steps between \emph{neutral} and the five used emotions. 

\bgroup
\newcolumntype{Y}{>{\centering\arraybackslash}X}
\newcolumntype{Z}{>{\raggedleft\arraybackslash}X}
\begin{table}[htb]
	\centering
	\begin{tabularx}{\columnwidth}{@{}XYYYY@{}}
		\toprule[\lightrulewidth]
		& \multicolumn{2}{c}{AffectNet Baseline} & \multicolumn{2}{c}{Evaluation Model}\\
		\cmidrule(l){2-3}
		\cmidrule(l){4-5}
		& Valence & Arousal & Valence &  Arousal \\
		\midrule[0.1em]
		RMSE & 0.37 & 0.41 & 0.40 & 0.37\\
		CORR & 0.66 & 0.54 & 0.60 & 0.52\\
		SAGR & 0.74 & 0.65 & 0.73 & 0.75\\
		CCC  & 0.60 & 0.34 & 0.57 & 0.44\\
		\bottomrule[\lightrulewidth]
		\addlinespace[\belowrulesep]
	\end{tabularx}
	\caption{AffectNet performance comparison. \cite{mertes2021continuous}}
	\label{tbl:affectnet-performance}
\end{table}
\egroup

The auxiliary classifier model was based on the MobileNetV2 architecture \citep{Sandler2018}. 
The model was trained on the AffectNet dataset for 100 epochs with \emph{Adam} optimizer and a learning rate of 0.001, leading to a similar performance as the AffectNet baseline models, as can be seen in Table \ref{tbl:affectnet-performance}.
We assessed the valence/arousal values for every interpolated output image of the cGAN and averaged them over the 1,000 samples per emotion, analogously to the feasibility studies described in Section \ref{sec:ceval}.
The results can be taken from Fig. \ref{fig:comp_eval}.



\section{Discussion}
In our initial study we evaluated if our proposed approach can be used to seamlessly interpolate images between generic classes. 
To this end we relied on two widely used and publicly available datasets CIFAR-10 and Fashion-MNIST (see \ref{sec:feasability_datasets}) to train our cGAN interpolation model.
Figure \ref{fig:fashion_morphs} and Figure \ref{fig:cifar_morphs} are showing examples of the calculated interpolations between various classes on the Fashion-MNIST dataset and the CIFAR-10 dataset respectively.
We can clearly see that the trained network was not able to capture the complexity of the input domain optimally. 
While the images from the Fashion-MNIST domain are showing slightly blurred contours, the generated images from the CIFAR-10 domain can only be partially recognized as the intended objects. 
However, when looking at the individual morphing steps between the classes, we can observe that the model is able to generate transitions that are generally smooth and continuous - two necessary prerequisites to apply the approach to interpolate between emotional expressions in human faces, in order to create meaningful results.

To further validate this observation we also employed a trained classifier for each dataset to predict the various interpolation steps between classes. 
Assuming a well calibrated classifier, we expected the distribution of the predicted class probabilities to continuously shift between the two interpolated classes along with the degree of interpolation. 

Figure \ref{fig:comp_eval_fashion} and Figure \ref{fig:comp_eval_cifar} are showing the results for those classifiers as described in Section \ref{sec:ceval}.
In those plots, we averaged the class probabilities for both the base classes and the target classes for every interpolation step of all the assessed output images.
It can be observed that, generally speaking, the interpolation mechanism led the network to generate transitions that are indeed perceived as \emph{lying between the classes} by the classifier. 
Notably, the images produces by the cGAN that was trained on CIFAR-10 were generally classified with quite low confidence. 
This implies that the assumption that the cGAN model was not able to accurately resemble the dataset holds true. 
However, for both models, the confidence smoothly transitions between the two intended classes, indicating that label interpolation is a promising tool for further experiments.

We argue that those findings further substantiate the ability of our trained cGAN to generate continuous interpolations between images and therefore also the feasibility to further investigate if the approach is able to generate meaningful interpolations between different categorical emotions.

\begin{figure}[]
	\centering
	\includegraphics[width=1\linewidth]{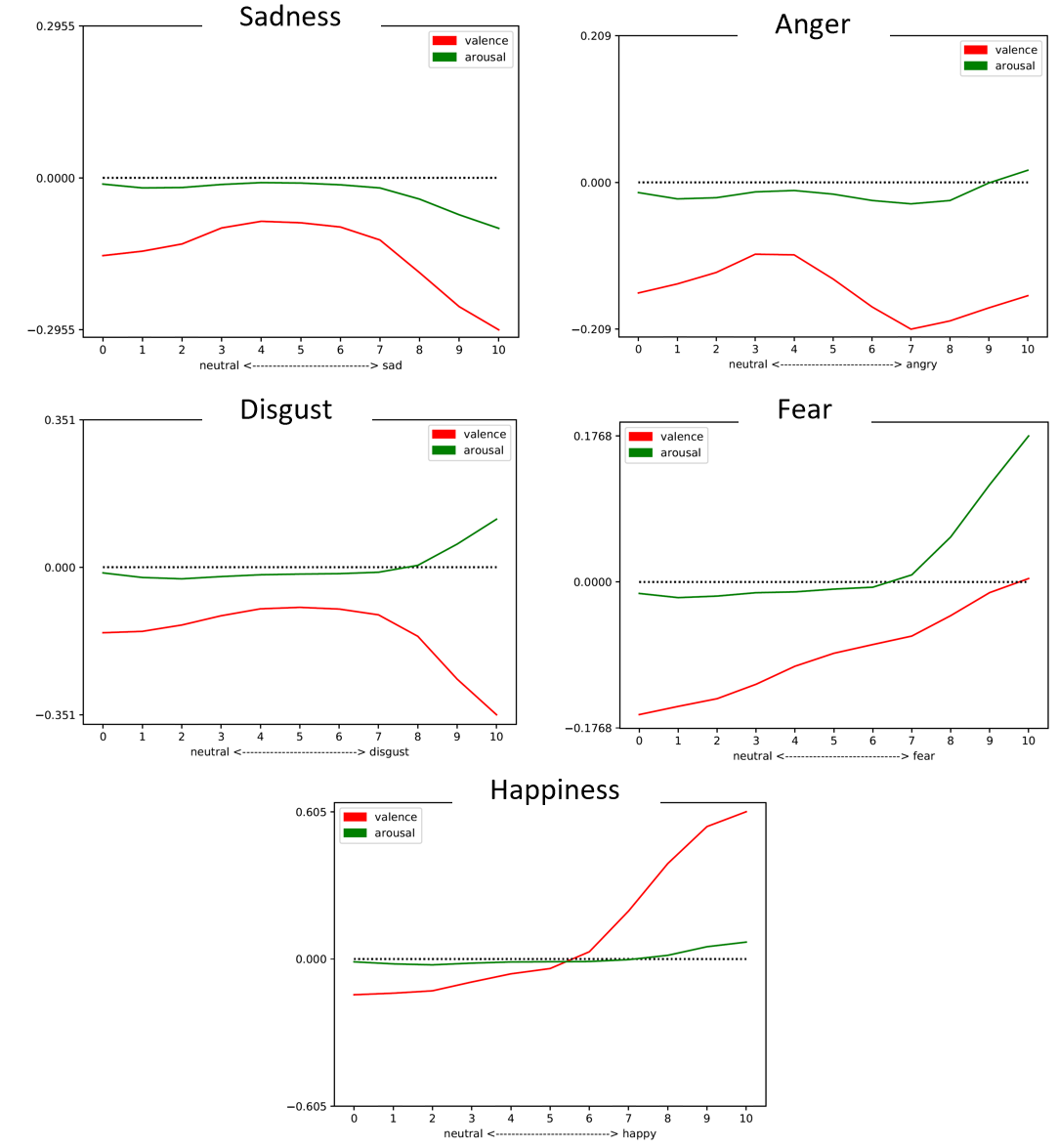}
	\caption{Computational Evaluation of our interpolation approach. Red graphs show valence, while green graphs show arousal. The x-axis represents the interpolation steps. Each interpolation step was performed by increasing the corresponding emotion vector element by 0.1, while decreasing the neutral vector element by 0.1. \cite{mertes2021continuous}}
	\label{fig:comp_eval}
\end{figure}

Upon visual inspection of the Fashion-MNIST and CIFAR-10 datasets, we found that the quality of the artificially generated images from the random noise vectors lagged significantly behind the original samples from the respective areas.

We therefore firstly conducted a user study to assess the capabilities of our employed cGAN model to produce realistic images of people expressing clearly identifiable emotions (see Sec. \ref{sec:user_eval}. 
The results of this study, as depicted in Fig. \ref{fig:eval_study}, show that participants generally recognize the expressed emotions in the artificially generated images similarly well as in the original images from the FACES dataset. 
The only exception being the emotion \emph{Sadness}, which was even better identifiable from the artificially generated images than the original data, where participants confused the emotion more often with \emph{Disgust}.
Those results are leading us to the conclusion that our employed cGAN model is suitable to further explore interpolation between emotional classes.

The results of the computational evaluation are depicted in Fig.\ref{fig:comp_eval} for each emotional class respectively. 
We can see that the interpolation mechanism is able to condition the cGAN to produce face images with various valence/arousal values. 
Upon further inspection we can observe that those values are mostly located in the value range between the start and end point of the interpolation, which indicates the general trend of the system to transition smoothly between emotional states. 
However, the plots also show that the interpolation function is not in all cases strictly monotonic. 
For example, in the \emph{Sadness} and \emph{Disgust} cases, the valence value initially rises slightly before dropping towards the interpolation endpoint. 
Similarly for \emph{Anger}, both valence and arousal values are first moving up and down before arriving at their initial starting point. 
This is a strong deviation to the position of anger in the circumplex model of emotions, where we would expect both valence and arousal to be notably higher when compare with the neutral position. 
Furthermore, we can see that the detected valence value is in all cases a bit below zero for the neural emotion.   
Since all emotions have been correctly recognized by human raters we attribute this behaviour to shortcomings in the valence arousal regressor. 
Taking those human quality ratings and the predominantly correct trend lines of the interpolation into account we argue that our approach can indeed be used to generate face images of continuous emotional states.
The fact, that the values are not evolving in a linear way, i.e., the plots appear rather as curves than as straight lines, does not take away much from the results, since the single interpolation step intervals can easily be modified to achieve a more even interpolation.
E.g., instead of using the same step interval for every single interpolation step, higher intervals can be used in ranges where the target features are changing slower.

\section{Conclusion \& Outlook}\label{sec_conclusion}
In this paper, we examined the possibilities of continuous interpolation through a discrete label space of Conditional Generative Adversarial Networks. Therefore, we first conducted some feasibility studies to assess the general applicability of interpolating between discrete classes to a trained cGAN. We found that indeed the technique can be used to generate smooth transitions between classes, even in cases where the cGAN did not learn to model the training domain to a satisfactory level.
Subsequently, we applied the label interpolation mechanism to the scenario of continuous emotional face generation. After ensuring that a cGAN trained on a dataset of categorical emotional face images learned to model that categorical emotional states by conducting a user study, we assessed the applicability of label interpolation in order to generate face images that show continuous emotional states. By using an auxiliary classifier for evaluating the cGAN outputs, we found that the algorithm was able to cover most of the valence/arousal ranges that are needed to cover the full dimensional emotion space. Although the performance of the approach shows to be highly dependent on the emotions that are used for interpolation, it shows great potential for application in various use cases such as automatic generation of virtual avatars or crowd generation.
In future work, it seems promising to apply label interpolation to GAN models with higher complexity in order to improve the quality of the generated results. Also, it is conceivable to use the proposed system for the task of data augmentation. Previous work has shown that GANs in general have the ability to enhance datasets in order to improve various deep learning tasks, such as semantic segmentation of images \citep{scherer2021unsupervised, choi2019self, mertes2020data, uricar2019let} or various image- and audio-based classification problems \citep{mariani2018bagan, frid2018synthetic, waheed2020covidgan, mertes2020evolutionary}. The ability to abstract from discrete classes to continuous features opens up a variety of machine learning problems where label interpolation could improve performance through data augmentation, which we plan to study in further research.

\section*{\uppercase{Acknowledgements}}
This work has been funded by the European Union Horizon 2020 research and innovation programme, grant agreement 856879.

\bibliographystyle{apalike}
{\small
\bibliography{main}}

\begin{thebibliography}{}

\bibitem[Arjovsky and Bottou, 2017]{arjovsky2017towards}
Arjovsky, M. and Bottou, L. (2017).
\newblock Towards principled methods for training generative adversarial
  networks.
\newblock {\em arXiv preprint arXiv:1701.04862}.

\bibitem[Arjovsky et~al., 2017]{arjovsky2017wasserstein}
Arjovsky, M., Chintala, S., and Bottou, L. (2017).
\newblock Wasserstein generative adversarial networks.
\newblock In {\em International conference on machine learning}, pages
  214--223. PMLR.

\bibitem[Beaupr{\'e} et~al., 2000]{msfde}
Beaupr{\'e}, M., Cheung, N., and Hess, U. (2000).
\newblock The montreal set of facial displays of emotion.
\newblock {\em Montreal, Quebec, Canada}.

\bibitem[Choi et~al., 2019]{choi2019self}
Choi, J., Kim, T., and Kim, C. (2019).
\newblock Self-ensembling with gan-based data augmentation for domain
  adaptation in semantic segmentation.
\newblock In {\em Proceedings of the IEEE/CVF International Conference on
  Computer Vision}, pages 6830--6840.

\bibitem[Choi et~al., 2018]{choi2018stargan}
Choi, Y., Choi, M., Kim, M., Ha, J.-W., Kim, S., and Choo, J. (2018).
\newblock Stargan: Unified generative adversarial networks for multi-domain
  image-to-image translation.
\newblock In {\em Proceedings of the IEEE conference on computer vision and
  pattern recognition}, pages 8789--8797.

\bibitem[Ding et~al., 2018]{ding2018exprgan}
Ding, H., Sricharan, K., and Chellappa, R. (2018).
\newblock Exprgan: Facial expression editing with controllable expression
  intensity.
\newblock In {\em Proceedings of the AAAI Conference on Artificial
  Intelligence}, volume~32.

\bibitem[Ebner et~al., 2010]{faces}
Ebner, N.~C., Riediger, M., and Lindenberger, U. (2010).
\newblock Faces—a database of facial expressions in young, middle-aged, and
  older women and men: Development and validation.
\newblock {\em Behavior research methods}, 42(1):351--362.

\bibitem[Frid-Adar et~al., 2018]{frid2018synthetic}
Frid-Adar, M., Klang, E., Amitai, M., Goldberger, J., and Greenspan, H. (2018).
\newblock Synthetic data augmentation using gan for improved liver lesion
  classification.
\newblock In {\em 2018 IEEE 15th international symposium on biomedical imaging
  (ISBI 2018)}, pages 289--293. IEEE.

\bibitem[Gauthier, 2014]{gauthier2014conditional}
Gauthier, J. (2014).
\newblock Conditional generative adversarial nets for convolutional face
  generation.
\newblock {\em Class Project for Stanford CS231N: Convolutional Neural Networks
  for Visual Recognition, Winter semester}, 2014(5):2.

\bibitem[Gong and Nass, 2007]{gong2007talking}
Gong, L. and Nass, C. (2007).
\newblock When a talking-face computer agent is half-human and half-humanoid:
  Human identity and consistency preference.
\newblock {\em Human communication research}, 33(2):163--193.

\bibitem[Goodfellow et~al., 2014]{goodfellow2014generative}
Goodfellow, I.~J., Pouget-Abadie, J., Mirza, M., Xu, B., Warde-Farley, D.,
  Ozair, S., Courville, A., and Bengio, Y. (2014).
\newblock Generative adversarial networks.
\newblock {\em arXiv preprint arXiv:1406.2661}.

\bibitem[Gulrajani et~al., 2017]{gulrajani2017improved}
Gulrajani, I., Ahmed, F., Arjovsky, M., Dumoulin, V., and Courville, A. (2017).
\newblock Improved training of wasserstein gans.
\newblock {\em arXiv preprint arXiv:1704.00028}.

\bibitem[He et~al., 2019]{he2019attgan}
He, Z., Zuo, W., Kan, M., Shan, S., and Chen, X. (2019).
\newblock Attgan: Facial attribute editing by only changing what you want.
\newblock {\em IEEE Transactions on Image Processing}, 28(11):5464--5478.

\bibitem[Karras et~al., 2017]{karras2017progressive}
Karras, T., Aila, T., Laine, S., and Lehtinen, J. (2017).
\newblock Progressive growing of gans for improved quality, stability, and
  variation.
\newblock {\em arXiv preprint arXiv:1710.10196}.

\bibitem[Kiderle et~al., 2021]{kiderle2021personality}
Kiderle, T., Ritschel, H., Janowski, K., Mertes, S., Lingenfelser, F., and
  Andr{\'e}, E. (2021).
\newblock Socially-aware personality adaptation.
\newblock In {\em 2021 9th International Conference on Affective Computing and
  Intelligent Interaction Workshops and Demos (ACIIW), in press}. IEEE.

\bibitem[Kossaifi et~al., 2017]{kossaifi2017afew}
Kossaifi, J., Tzimiropoulos, G., Todorovic, S., and Pantic, M. (2017).
\newblock Afew-va database for valence and arousal estimation in-the-wild.
\newblock {\em Image and Vision Computing}, 65:23--36.

\bibitem[Krizhevsky and Hinton, 2009]{krizhevsky2009learning}
Krizhevsky, A. and Hinton, G. (2009).
\newblock Learning multiple layers of features from tiny images (technical
  report).
\newblock {\em University of Toronto}.

\bibitem[Lang et~al., 1997a]{Lang1997}
Lang, P.~J., Bradley, M.~M., and Cuthbert, B.~N. (1997a).
\newblock Motivated attention: Affect, activation, and action.
\newblock In Lang, P.~J., Simons, R.~F., and Balaban, M.~T., editors, {\em
  Attention and orienting: Sensory and motivational processes}, pages 97--135.
  Psychology Press.

\bibitem[Lang et~al., 1997b]{iaps}
Lang, P.~J., Bradley, M.~M., Cuthbert, B.~N., et~al. (1997b).
\newblock International affective picture system (iaps): Technical manual and
  affective ratings.
\newblock {\em NIMH Center for the Study of Emotion and Attention}, 1:39--58.

\bibitem[Lin et~al., 2018]{lin2018conditional}
Lin, J., Xia, Y., Qin, T., Chen, Z., and Liu, T.-Y. (2018).
\newblock Conditional image-to-image translation.
\newblock In {\em Proceedings of the IEEE conference on computer vision and
  pattern recognition}, pages 5524--5532.

\bibitem[Liu et~al., 2017]{liu2017unsupervised}
Liu, M.-Y., Breuel, T., and Kautz, J. (2017).
\newblock Unsupervised image-to-image translation networks.
\newblock {\em arXiv preprint arXiv:1703.00848}.

\bibitem[Lucey et~al., 2010]{ck+}
Lucey, P., Cohn, J.~F., Kanade, T., Saragih, J., Ambadar, Z., and Matthews, I.
  (2010).
\newblock The extended cohn-kanade dataset (ck+): A complete dataset for action
  unit and emotion-specified expression.
\newblock In {\em 2010 ieee computer society conference on computer vision and
  pattern recognition-workshops}, pages 94--101. IEEE.

\bibitem[Mariani et~al., 2018]{mariani2018bagan}
Mariani, G., Scheidegger, F., Istrate, R., Bekas, C., and Malossi, C. (2018).
\newblock Bagan: Data augmentation with balancing gan.
\newblock {\em arXiv preprint arXiv:1803.09655}.

\bibitem[Matsumoto, 1988]{jacfee}
Matsumoto, D.~R. (1988).
\newblock {\em Japanese and Caucasian facial expressions of emotion (JACFEE)}.
\newblock University of California.

\bibitem[Mehrabian, 1995]{Mehrabian1995}
Mehrabian, A. (1995).
\newblock {Framework for a comprehensive description and measurement of
  emotional states.}
\newblock {\em Genetic, social, and general psychology monographs},
  121(3):339--361.

\bibitem[Mertes et~al., 2020a]{mertes2020evolutionary}
Mertes, S., Baird, A., Schiller, D., Schuller, B.~W., and Andr{\'e}, E.
  (2020a).
\newblock An evolutionary-based generative approach for audio data
  augmentation.
\newblock In {\em 2020 IEEE 22nd International Workshop on Multimedia Signal
  Processing (MMSP)}, pages 1--6. IEEE.

\bibitem[Mertes et~al., 2021a]{mertes2021voice}
Mertes, S., Kiderle, T., Schlagowski, R., Lingenfelser, F., and Andr{\'e}, E.
  (2021a).
\newblock On the potential of modular voice conversion for virtual agents.
\newblock In {\em 2021 9th International Conference on Affective Computing and
  Intelligent Interaction Workshops and Demos (ACIIW), in press}. IEEE.

\bibitem[Mertes et~al., 2021b]{mertes2021continuous}
Mertes, S., Lingenfelser, F., Kiderle, T., Dietz, M., Diab, L., and
  Andr{\'{e}}, E. (2021b).
\newblock Continuous emotions: Exploring label interpolation in conditional
  generative adversarial networks for face generation.
\newblock In Fred, A. L.~N., Sansone, C., and Madani, K., editors, {\em
  Proceedings of the 2nd International Conference on Deep Learning Theory and
  Applications, DeLTA 2021, Online Streaming, July 7-9, 2021}, pages 132--139.
  {SCITEPRESS}.

\bibitem[Mertes et~al., 2020b]{mertes2020data}
Mertes, S., Margraf, A., Kommer, C., Geinitz, S., and Andr{\'{e}}, E. (2020b).
\newblock Data augmentation for semantic segmentation in the context of carbon
  fiber defect detection using adversarial learning.
\newblock In Fred, A. L.~N. and Madani, K., editors, {\em Proceedings of the
  1st International Conference on Deep Learning Theory and Applications, DeLTA
  2020, Lieusaint, Paris, France, July 8-10, 2020}, pages 59--67. ScitePress.

\bibitem[Mirza and Osindero, 2014]{mirza2014conditional}
Mirza, M. and Osindero, S. (2014).
\newblock Conditional generative adversarial nets.
\newblock {\em arXiv preprint arXiv:1411.1784}.

\bibitem[Mollahosseini et~al., 2017]{mollahosseini2017affectnet}
Mollahosseini, A., Hasani, B., and Mahoor, M.~H. (2017).
\newblock Affectnet: A database for facial expression, valence, and arousal
  computing in the wild.
\newblock {\em IEEE Transactions on Affective Computing}, 10(1):18--31.

\bibitem[Radford et~al., 2015]{radford2015unsupervised}
Radford, A., Metz, L., and Chintala, S. (2015).
\newblock Unsupervised representation learning with deep convolutional
  generative adversarial networks.
\newblock {\em arXiv preprint arXiv:1511.06434}.

\bibitem[Ritschel et~al., 2019]{ritschel2019personalized}
Ritschel, H., Aslan, I., Mertes, S., Seiderer, A., and Andr{\'e}, E. (2019).
\newblock Personalized synthesis of intentional and emotional non-verbal sounds
  for social robots.
\newblock In {\em 2019 8th International Conference on Affective Computing and
  Intelligent Interaction (ACII)}, pages 1--7. IEEE.

\bibitem[Royer et~al., 2020]{royer2020xgan}
Royer, A., Bousmalis, K., Gouws, S., Bertsch, F., Mosseri, I., Cole, F., and
  Murphy, K. (2020).
\newblock Xgan: Unsupervised image-to-image translation for many-to-many
  mappings.
\newblock In {\em Domain Adaptation for Visual Understanding}, pages 33--49.
  Springer.

\bibitem[Russell and Barrett, 1999]{russell1999core}
Russell, J.~A. and Barrett, L.~F. (1999).
\newblock Core affect, prototypical emotional episodes, and other things called
  emotion: dissecting the elephant.
\newblock {\em Journal of personality and social psychology}, 76(5):805.

\bibitem[{Sandler} et~al., 2018]{Sandler2018}
{Sandler}, M., {Howard}, A., {Zhu}, M., {Zhmoginov}, A., and {Chen}, L. (2018).
\newblock Mobilenetv2: Inverted residuals and linear bottlenecks.
\newblock In {\em 2018 IEEE/CVF Conference on Computer Vision and Pattern
  Recognition}, pages 4510--4520.

\bibitem[Scherer et~al., 2021]{scherer2021unsupervised}
Scherer, S., Sch{\"o}n, R., Ludwig, K., and Lienhart, R. (2021).
\newblock Unsupervised domain extension for nighttime semantic segmentation in
  urban scenes.

\bibitem[Schlagowski et~al., 2021]{schlagowski2021taming}
Schlagowski, R., Mertes, S., and Andr{\'e}, E. (2021).
\newblock Taming the chaos: exploring graphical input vector manipulation user
  interfaces for gans in a musical context.
\newblock In {\em Audio Mostly 2021}, pages 216--223.

\bibitem[Schrum et~al., 2020]{schrum2020interactive}
Schrum, J., Gutierrez, J., Volz, V., Liu, J., Lucas, S., and Risi, S. (2020).
\newblock Interactive evolution and exploration within latent level-design
  space of generative adversarial networks.
\newblock In {\em Proceedings of the 2020 Genetic and Evolutionary Computation
  Conference}, pages 148--156.

\bibitem[Tan and Le, 2019]{tan2019efficientnet}
Tan, M. and Le, Q. (2019).
\newblock Efficientnet: Rethinking model scaling for convolutional neural
  networks.
\newblock In {\em International Conference on Machine Learning}, pages
  6105--6114. PMLR.

\bibitem[Tottenham, 1998]{nimstim}
Tottenham, N. (1998).
\newblock Macbrain face stimulus set.
\newblock {\em John D. and Catherine T. MacArthur Foundation Research Network
  on Early Experience and Brain Development}.

\bibitem[Uricar et~al., 2019]{uricar2019let}
Uricar, M., Sistu, G., Rashed, H., Vobecky, A., Kumar, V.~R., Krizek, P.,
  Burger, F., and Yogamani, S. (2019).
\newblock Let's get dirty: Gan based data augmentation for camera lens soiling
  detection in autonomous driving.
\newblock {\em arXiv preprint arXiv:1912.02249}.

\bibitem[Van~der Schalk et~al., 2009]{adfes}
Van~der Schalk, J., Hawk, S., and Fischer, A. (2009).
\newblock Validation of the amsterdam dynamic facial expression set (adfes).
\newblock {\em Poster for the International Society for Research on Emotions
  (ISRE), Leuven, Belgium}.

\bibitem[van Rijn et~al., 2021]{rijn21_interspeech}
van Rijn, P., Mertes, S., Schiller, D., Harrison, P.~M., Larrouy-Maestri, P.,
  André, E., and Jacoby, N. (2021).
\newblock {Exploring Emotional Prototypes in a High Dimensional TTS Latent
  Space}.
\newblock In {\em Proc. Interspeech 2021}, pages 3870--3874.

\bibitem[Waheed et~al., 2020]{waheed2020covidgan}
Waheed, A., Goyal, M., Gupta, D., Khanna, A., Al-Turjman, F., and Pinheiro,
  P.~R. (2020).
\newblock Covidgan: data augmentation using auxiliary classifier gan for
  improved covid-19 detection.
\newblock {\em Ieee Access}, 8:91916--91923.

\bibitem[Wang et~al., 2018]{wang2018attributes}
Wang, Y., Dantcheva, A., and Bremond, F. (2018).
\newblock From attributes to faces: a conditional generative network for face
  generation.
\newblock In {\em 2018 International Conference of the Biometrics Special
  Interest Group (BIOSIG)}, pages 1--5. IEEE.

\bibitem[Xiao et~al., 2017]{fashion-mnist}
Xiao, H., Rasul, K., and Vollgraf, R. (2017).
\newblock Fashion-mnist: a novel image dataset for benchmarking machine
  learning algorithms.
\newblock {\em CoRR}, abs/1708.07747.

\bibitem[Yi et~al., 2018]{yi2018data}
Yi, W., Sun, Y., and He, S. (2018).
\newblock Data augmentation using conditional gans for facial emotion
  recognition.
\newblock In {\em 2018 Progress in Electromagnetics Research Symposium
  (PIERS-Toyama)}, pages 710--714. IEEE.

\end{thebibliography}

\end{document}